\pdfoutput=1

\documentclass[11pt]{article}

\usepackage{emnlp2021}

\usepackage{times}
\usepackage{latexsym}
\usepackage{multirow}
\usepackage{todonotes}
\usepackage{colortbl}
\usepackage[T1]{fontenc}

\usepackage[utf8]{inputenc}

\usepackage{microtype}

\usepackage{graphicx}

\usepackage{xspace}
\usepackage{ amssymb }

\newcount\numNegADETrain
\newcount\numNoADETrain
\newcount\numADETrain
\newcount\numNegADETest
\newcount\numNoADETest
\newcount\numADETest

\newcount\totTrain
\newcount\totTest

\newcommand{\smallsf}[1]{\textsf{\small #1}}
\newcommand{\smalltt}[1]{\textbf{\texttt{\small #1}}}

\newcommand{\ade}{\smallsf{ADE}\xspace}
\newcommand{\noade}{\smallsf{noADE}\xspace}
\newcommand{\negade}{\smallsf{negADE}\xspace}
\newcommand{\negadeR}{\smallsf{negADE$_\texttt{\tiny R}$}\xspace}
\newcommand{\negadeG}{\smallsf{negADE$_\texttt{\tiny G}$}\xspace}

\renewcommand{\ade}{\smalltt{ADE}\xspace}
\renewcommand{\noade}{\smalltt{noADE}\xspace}
\renewcommand{\negade}{\smalltt{negADE}\xspace}
\renewcommand{\negadeR}{\smalltt{negADE\_r}\xspace}
\renewcommand{\negadeG}{\smalltt{negADE\_g}\xspace}

\newcommand{\SMMnum}[1]{S{\small MM4H#1}\xspace}
\newcommand{\SMM}{\SMMnum{}}
\newcommand{\SMMi}{\SMMnum{19$_{ext}$}}

\newcommand{\SMMib}{\SMMnum{19$_{cls}$}}
\newcommand{\SMMiib}{\SMMnum{20$_{cls}$}}

\newcommand{\bertneg}{BERTneg\xspace}

\newcommand{\datasetname}{NADE\xspace}

\newcommand{\cellred}{\cellcolor{gray!75}}
\newcommand{\cellgreen}{\cellcolor{teal!30}}
\newcommand{\cellblue}{\cellcolor{orange!40}}

\newcommand{\colnamesIII}{
\multicolumn{1}{c}{\textbf{FP}} &
\multicolumn{3}{c}{\ade \noade \negade}
}

\makeatletter
\newcommand{\thickcline}[1]{%
    \@thickcline #1\@nil%
}

\def\@thickcline#1-#2\@nil{%
  \omit
  \@multicnt#1%
  \advance\@multispan\m@ne
  \ifnum\@multicnt=\@ne\@firstofone{&\omit}\fi
  \@multicnt#2%
  \advance\@multicnt-#1%
  \advance\@multispan\@ne
  \leaders\hrule\@height1pt\hfill
  \cr
  \noalign{\vskip-1pt}%
}
\makeatother
\newcommand{\mycell}[2]{
\the#1\hskip .2cm {\small
\the\numexpr 100*#1/#2\%}
}

\title{\datasetname: A Benchmark for Robust Adverse Drug Events Extraction \\ in Face of Negations}

\author{
Simone Scaboro$^{1}$ $\quad$ 
Beatrice Portelli$^1$ $\quad$  
\textbf{Emmanuele Chersoni}$^2$ \\
\textbf{Enrico Santus}$^3$ $\quad$  
\textbf{Giuseppe Serra}$^1$ \smallskip \\ 
  $^1$ University of Udine, Italy \\
  $^2$ The Hong Kong Polytechnic University, Hong Kong \\
  $^3$ DSIG - Bayer Pharmaceuticals, New Jersey, USA \smallskip \\
  \texttt{\{scaboro.simone,portelli.beatrice\}@spes.uniud.it},\\
  
  \texttt{emmanuele.chersoni@polyu.edu.hk},\\
  \texttt{esantus@gmail.com},\hskip .25cm \texttt{giuseppe.serra@uniud.it}
  }

\begin{document}

\totTrain = \numexpr \numADETrain+\numNoADETrain+\numNegADETrain
\totTest = \numexpr \numADETest+\numNoADETest+\numNegADETest

\maketitle
\begin{abstract}
Adverse Drug Event (ADE) extraction models can rapidly examine large collections of social media texts, detecting mentions of drug-related adverse reactions and trigger medical investigations.
However, despite the recent advances in NLP, it is currently unknown if such models are robust in face of \textit{negation}, which is pervasive across language varieties.

In this paper we evaluate three state-of-the-art systems,
showing their fragility against negation, and then we introduce two possible strategies to increase the robustness of these models: a pipeline approach, relying on a specific component for negation detection; an augmentation of an ADE extraction dataset to  artificially create negated samples and further train the models.

We show that both strategies bring significant increases in performance, lowering the number of spurious entities predicted by the models.
Our dataset and code will be publicly released to encourage research on the topic.

\end{abstract}

\section{Introduction}

Exploring social media texts is becoming more and more important in the field of pharmacovigilance
\citep{karimi2015text,Sarker2015PortableAT}
, since it is common for Internet users to report their personal experiences with drugs on forums and microblogging platforms.
Given the inherent noisiness of social media texts (colloquial language, slang and metaphors, non-standard syntactic constructions etc.), the Natural Language Processing (NLP) community dedicated a consistent effort in developing robust methods for mining biomedical information from social media outlets. This also led to the creation of several dedicated shared tasks series on ADE detection (\SMM\ -- Social Media Mining for Health) \citep{paul2016overview,sarker2017overview,weissenbacher2018overview,weissenbacher2019overview,klein2020overview}.

Although these models have seen great advancements in the last years, also thanks to the introduction of pre-trained Transformers-based architectures \citep{vaswani2017attention,Devlin2019BERTPO}, it is still unknown how robust they are in face of some pervasive linguistic phenomena such as negation.
However, general investigations on machine comprehension and question answering tasks confirmed that such phenomena often pose a serious challenge \cite{ribeiro2020beyond}.
Managing to efficiently handle the scope of negations and speculations
in clinical notes is a key problem in biomedical NLP \citep{velldal2012speculation,diaz2013detecting}, and similarly, for digital pharmacovigilance it is essential to recognize whether the association between a drug and an ADE is actually being stated or negated because the consequences of extracting misleading information about the possible side effects of drugs can be extremely serious. 

In this paper, we analyze the performance of some of the latest state-of-the-art ADE detection systems on NADE:
a new dataset derived from \SMM data, which contains a relevant amount of annotated samples with negated ADEs.
We also introduce and analyze two strategies to increase the robustness of the models: adding a negation detection module in a pipeline fashion to exclude the negated ADEs predicted by the models; augmenting the training set with artificially negated samples.
As a further contribution, our dataset and scripts will be made publicly available for researchers to test the robustness of their ADE extraction systems against negation.\footnote{\url{https://github.com/AilabUdineGit/NADE-dataset}}

\section{Related Work}

Detecting negation scopes is a traditional topic of NLP research.
An early, popular system was introduced by \citet{chapman2001simple}, whose NegEx algorithm exploited regular expressions to identify negations in clinical documents in English.
Later, machine learning approaches became more popular after the publication of a common gold standard, i.e.~the BioScope corpus \citep{vincze2008bioscope}. Several proposals consisted of a two-steps methodology: a first classifier to detect which token in a sentence is a negation/speculation cue, and a second classifier to determine which tokens in the sentence are within the scope of the cue word \citep{morante2008learning,cruz2012machine,attardi2015detecting,zou2015negation}.

More recently, approaches based on neural networks (CNN, \citealt{qian2016speculation}; BiLSTM, \citealt{fancellu2016neural,fancellu2017detecting,dalloux2019speculation}) have been introduced in the literature, showing some crosslinguistic and crossdomain transferability.
Moreover, BERT-based models have been proposed to handle this phenomenon \citep{khandelwal2019negbert,britto2020resolving}, also with the aid of multitask learning architectures \citep{khandelwal2020multitask}.

To our knowledge, the research in biomedical NLP mostly focused on scope detection \textit{per se} and on more formal types of texts (e.g. clinical notes, articles). In our research we focus instead on the specific task of ADE detection and on the impact of negation on the performance of ADE systems for noisy social media texts (e.g. tweets, blog posts), with the goal of making them able to distinguish between factual and non-factual information.

\section{\datasetname Dataset}

While there are several datasets for ADE detection on social media texts \citep{Karimi2015CadecAC, twimed}, the biggest collection of tweets tagged for ADE mentions
is the one released yearly for the \SMM Workshop and Shared Task.
We used the following resources:

a) \textbf{\SMMi}. The training set for the ADE \textit{extraction} Task of \SMMnum{19} \citep{weissenbacher2019overview}, consisting of 2276 tweets that mention at least one drug name. 1300 of them contain ADEs, and annotations of their position in the text (\ade class). The other 976 are control samples with no ADE mentions (\noade).
As the blind test set is not publicly available, we rely on  
the training set only and use the splits by \citet{portelli2021bert}, which balance positive and negative samples;

b) \textbf{\SMMib} and \textbf{\SMMiib} \citep{weissenbacher2019overview,klein2020overview}. The training sets for \SMM \textit{classification} Tasks, containing tweets labeled as \ade or \noade (1:9 ratio).

The community focused on the ADE extraction task, so most datasets are made of samples that either do or do not contain an ADE. Because of this, they include a small number of negated ADEs by construction: no particular attention is given to these samples when curating the data and when they are present they are treated as \noade samples and not explicitly labelled. This leads to their class being misrepresented and makes it harder to study this phenomenon.

\subsection{Data Augmentation}

In order to perform our analysis we
created a new set of samples containing negated ADEs (\negade) in two ways:
looking for real samples negating the presence of an ADE in \SMMib and \SMMiib (\negadeR);
manually creating negated versions for the \ade tweets in the test split of \SMMi (\negadeG).
The original test split includes 260 tweets, 7 of which were discarded during the generation process, leading to 253 samples. Further details in Appendix A.

\subsubsection*{Recovery of Real Samples}
\SMMib and \SMMiib contain a total of 24857 unique \noade tweets, so there is an high chance of encountering negated ADEs. The samples have no other annotation apart from their binary labels and analyzing all of them manually would be extremely time-consuming.
We performed a preliminary filtering, keeping only the tweets containing a negation cue.
Then we manually analyzed the filtered tweets, assessing whether the negation refers to an ADE and if the message actually negates the presence of the ADE.
In the following examples: the first tweet is valid (it negates the ADE); the second one contains a negation, but does not negate the ADE. 

\smallsf{\textbf{1.} This \textbf{\#HUMIRA shot} has me feeling like a normal}

\smallsf{human... \textbf{No pain no inflammation} no nothinggggh}

\smallsf{\textbf{2.}  But I'm \textbf{not on adderall and I am feasting}.}
\smallskip

The tweets were evaluated by four volunteer annotators with a high level of proficiency in English, and we only kept the samples for which they were in agreement. As a result we obtained \negadeR, a set of real tweets containing negated ADEs, that allows us to create a test set containing up to 16\% negated samples.

\subsubsection*{Generation of Artificial Samples}
The generation process for the \negadeG samples was carried out by the same four volunteers as before. Each one of them was given part of the 260 tweets from the \SMMi test set, and was asked to alter them (as conservatively as possible) to generate a new version of the tweet that negates the presence of the ADE. Each volunteer was asked to review the tweets generated by the other participants and to propose modifications, in case the tweets looked ambiguous or unnatural. If no agreement was found about the edits, the augmented tweets were discarded.
The result of this procedure is \negadeG, a new set of tweets denying the presence of an ADE. Here is an example of an original tweet and its negated version (highlighting the cue word added to negate the ADE):

\smallsf{\textbf{Original}: fluoxetine, got me going crazy.}

\smallsf{\textbf{Negated}: fluoxetine, \textbf{didn't} get me going crazy.}

\subsection{Data Partitioning}

We split the available data in a train and a test set, both containing the three categories of tweets: \ade, \noade and \negade (Table \ref{tab:dataset_composition}).
Given the small amount of \negadeR tweets, we use all of them in the test set to evaluate the performance only on real tweets.
Conversely, the training set only contains the manually generated \negadeG samples.

\begin{table}[htbp]%
\centering%
\resizebox{.9\linewidth}{!}{%
\begin{tabular}{r*{3}{r}r}

\multicolumn{1}{c}{}&

\multicolumn{1}{c}{\textbf{\ade}} &
\multicolumn{1}{c}{\textbf{\noade}} &
\multicolumn{1}{c}{\textbf{\negade}} &
\multicolumn{1}{c}{\textbf{{Total}}} \\ 

\hline

\textbf{{Train}} &
\mycell{\numADETrain}{\totTrain} &
\mycell{\numNoADETrain}{\totTrain} &
\mycell{\numNegADETrain}{\totTrain} &
\the\totTrain
\\
\textbf{{Test}} &
\mycell{\numADETest}{\totTest} &
\mycell{\numNoADETest}{\totTest} &
\mycell{\numNegADETest}{\totTest} &
\the\totTest
\\
\hline%
\end{tabular}%
}%
\caption{Distribution of samples in \datasetname.}
\label{tab:dataset_composition}
\end{table}

\section{Analyzed models}

\subsection{ADE Extraction Models}

We choose three Transformer-based models that showed high performance on \SMMi \citep{portelli2021bert,portelli2021improving}, and are currently at the top of the \SMMi ADE extraction leaderboard:
{BERT} 
\citep{Devlin2019BERTPO},
{SpanBERT} 
\citep{Joshi2019SpanBERTIP}
and {PubMedBERT} \citep{pubmedbert}.
The models are fine-tuned for token classification, predicting an IOB label for each token in the sentence to detect the boundaries of ADE mentions.

\subsection{Negation Detection Models}

We introduce two negation detection modules:
{NegEx}, a Python implementation \citep{jeno_pizarro_2020_4279555} of the NegEx algorithm, based on simple regular expressions, which evaluates whether named entities are negated;
{\bertneg}, a BERT model (bert-base-uncased) that we finetuned for token classification. We trained \bertneg on the BioScope dataset,
which consists in medical texts annotated for the presence of negation and speculation cues and their related scopes. We selected 3190 sentences (2801 of which with a negation scope) and finetuned the model for scope detection (10 epochs, learning rate $0.0001$).

\subsection{Pipeline Models}
Let us consider a text $t$, a ADE extraction base model $\mathcal{B}$ and a negation detection module $\mathcal{N}$.
Given $t$, $\mathcal{B}$ outputs a set of substrings of $t$ that are labeled as ADE mentions:
$\mathcal{B}(t) = \{b_1, \dots, b_m\}$.
Similarly, $\mathcal{N}$ takes a text and outputs a set of substrings, which are considered to be entities within a negation scope:
$\mathcal{N}(t) = \{n_1, \dots, n_t\}$.

A combined \textit{pipeline model} is obtained by discarding all ADE spans $b_i \in \mathcal{B}(t)$ that overlap one of the negation spans $n_j\in\mathcal{N}(t)$:\\
{\small
$\mathcal{B}\mathcal{N}(t)=\{b_i\in\mathcal{B}(t)\mid \forall j (n_j \in \mathcal{N}(t)\wedge b_i\cap n_j= \emptyset)\}$}

\section{Experiments}

\begin{table*}[htbp!]
\centering
\resizebox{!}{6em}{%

\begin{tabular}{rr@{\ \ }c@{\ \ }rc@{\ \ \ }c@{\ \ \ }l}

&&& \multicolumn{4}{c}{\textbf{ BERT}} \\
&&& \colnamesIII \\ \thickcline{4-7}

1& \textbf{$\mathcal{B}$ (base model)}           && 161.2 & \color{gray} 46.2 & \color{gray} 42.6 & \color{gray} 73.4  
\\ \cline{2-2} \cline{4-7}
2& \textbf{$\mathcal{B}$+NegEx}                  && 106.4 & \color{gray} 41.6 & \color{gray} 40.4 & \color{gray} 24.4  
\\
3& \textbf{$\mathcal{B}$+BERTneg}                && 120.2 & \color{gray} 40.0 & \color{gray} 41.0 & \color{gray} 39.2  
\\ \cline{2-2} \cline{4-7}
4& \textbf{$\mathcal{B}$\smalltt{+50}}           && 146.6 & \color{gray} 40.2 & \color{gray} 38.4 & \color{gray} 69.0  
\\
5& \textbf{$\mathcal{B}$\smalltt{+100}}          && 166.8 & \color{gray} 50.4 & \color{gray} 48.4 & \color{gray} 69.0  
\\
6& \textbf{$\mathcal{B}$\smalltt{+150}}          && 125.8 & \color{gray} 43.0 & \color{gray} 41.4 & \color{gray} 42.4  
\\
7& \textbf{$\mathcal{B}$\smalltt{+200}}          && 105.4 & \color{gray} 41.0 & \color{gray} 36.0 & \color{gray} 29.0  
\\
8& \textbf{$\mathcal{B}$\smalltt{+253}}          && 101.6 & \color{gray} 42.6 & \color{gray} 35.8 & \color{gray} 23.2  
\\ \cline{2-2} \cline{4-7}
9& \textbf{$\mathcal{B}$+NegEx} \smalltt{+253}   && 91.6  & \color{gray} 42.2 & \color{gray} 35.8 & \color{gray} 13.6  
\\
10& \textbf{$\mathcal{B}$+BERTneg} \smalltt{+253} && 93.6  & \color{gray} 41.0 & \color{gray} 35.8 & \color{gray} 16.6  
\\  \cline{2-2} \cline{4-7}

\end{tabular}%
\begin{tabular}{r@{}r@{}c@{\ \ }rc@{\ \ \ }c@{\ \ \ }l}

&&& \multicolumn{4}{c}{\textbf{ PubMedBERT}} \\
&&& \colnamesIII \\ \thickcline{4-7}

&&& 144.2 & \color{gray} 37.0 & \color{gray} 40.4 & \color{gray} 67.4  
\\ \cline{4-7} 
&&& 93.4  & \color{gray} 32.0 & \color{gray} 37.6 & \color{gray} 23.8  
\\
&&& 106.2 & \color{gray} 30.6 & \color{gray} 39.0 & \color{gray} 36.6  
\\ \cline{4-7}
&&& 126.2 & \color{gray} 34.4 & \color{gray} 35.4 & \color{gray} 57.0  
\\
&&& 108.0 & \color{gray} 33.0 & \color{gray} 34.6 & \color{gray} 40.4  
\\
&&& 100.8 & \color{gray} 29.8 & \color{gray} 41.4 & \color{gray} 29.6  
\\
&&& 79.0  & \color{gray} 29.2 & \color{gray} 27.6 & \color{gray} 22.2  
\\
&&& 84.2  & \color{gray} 30.6 & \color{gray} 34.6 & \color{gray} 19.0  
\\ \cline{4-7}
&&& 76.6  & \color{gray} 29.4 & \color{gray} 34.6 & \color{gray} 12.6  
\\
&&& 74.2  & \color{gray} 27.2 & \color{gray} 34.2 & \color{gray} 12.8  
\\ \cline{4-7}

\end{tabular}%
\begin{tabular}{r@{}r@{}c@{\ \ }rc@{\ \ \ }c@{\ \ \ }l}

&&& \multicolumn{4}{c}{\textbf{ SpanBERT}} \\
&&& \colnamesIII \\ \thickcline{4-7}

&&& 245.6 & \color{gray} 66.2 & \color{gray} 79.8 & \color{gray} 100.8 
\\ \cline{4-7}
&&& 170.8 & \color{gray} 58.4 & \color{gray} 74.0 & \color{gray} 38.6  
\\
&&& 184.0 & \color{gray} 56.2 & \color{gray} 74.8 & \color{gray} 53.2  
\\ \cline{4-7}
&&& 183.2 & \color{gray} 46.2 & \color{gray} 58.6 & \color{gray} 79.2  
\\
&&& 230.0 & \color{gray} 58.6 & \color{gray} 77.4 & \color{gray} 95.4  
\\
&&& 156.8 & \color{gray} 47.6 & \color{gray} 50.8 & \color{gray} 59.2  
\\
&&& 134.2 & \color{gray} 38.6 & \color{gray} 44.4 & \color{gray} 51.6  
\\
&&& 179.8 & \color{gray} 55.2 & \color{gray} 60.6 & \color{gray} 65.2  
\\ \cline{4-7}
&&& 136.4 & \color{gray} 52.0 & \color{gray} 57.0 & \color{gray} 27.6  
\\
&&& 145.4 & \color{gray} 51.4 & \color{gray} 57.4 & \color{gray} 36.8  
\\ \cline{4-7}

\end{tabular}
}
\caption{%
False Positives for: the base models; the pipeline models; base models trained with an increasing number of \negadeG samples; pipeline models trained with all \negadeG samples.
}
\label{tab:fp}
\end{table*}

\begin{table*}[htbp!]
\centering
\resizebox{!}{4.6em}{%

\begin{tabular}{rr@{\ \ }rc@{\ \ }c@{\ \ \ }c@{\ \ \ }}

&&& \multicolumn{3}{c}{\textbf{ BERT}} \\
&&& \textbf{P} & \textbf{R} & \textbf{F1} \\ \thickcline{4-6}

1& \textbf{$\mathcal{B}$ (base model)}           && 50.15 & 65.6 & 56.78    \\ \cline{2-2} \cline{4-6}
2& \textbf{$\mathcal{B}$+NegEx}                  && 55.59 & 58.03 & 56.73   \\
3& \textbf{$\mathcal{B}$+BERTneg}                && 54.37 & 60.7 & 57.30   \\ \cline{2-2} \cline{4-6}
$\vdots$&&$\ \vdots$\\
8& \textbf{$\mathcal{B}$\smalltt{+253}}          && 58.85 & 63.86 & 61.21  \\ \cline{2-2} \cline{4-6}
9& \textbf{$\mathcal{B}$+NegEx} \smalltt{+253}   && 58.65 & 58.03 & 58.29  \\
10& \textbf{$\mathcal{B}$+BERTneg} \smalltt{+253} && 59.22 & 60.17 & 59.65 \\  \cline{2-2} \cline{4-6}

\end{tabular}%
\begin{tabular}{r@{}r@{}rc@{\ \ }c@{\ \ \ }c@{\ \ \ }}

&&& \multicolumn{3}{c}{\textbf{ PubMedBERT}} \\
&&& \textbf{P} & \textbf{R} & \textbf{F1} \\ \thickcline{4-6}

&&& 53.24 & 67.41 & 59.47   \\ \cline{4-6} 
&&& 59.21 & 59.76 & 59.47  \\
&&& 57.69 & 62.64 & 60.04  \\ \cline{4-6}
$\vdots$\\
&&& 63.28 & 63.33 & 63.20  \\ \cline{4-6}
&&& 63.24 & 58.29 & 60.58  \\
&&& 64.74 & 60.98 & 62.72  \\ \cline{4-6}

\end{tabular}%
\begin{tabular}{r@{}r@{}rc@{\ \ }c@{\ \ \ }c@{\ \ \ }}

&&& \multicolumn{3}{c}{\textbf{ SpanBERT}} \\
&&& \textbf{P} & \textbf{R} & \textbf{F1} \\ \thickcline{4-6}

&&& 43.65 & 73.28 & 54.61  \\ \cline{4-6}
&&& 48.65 & 65.04 & 55.55  \\
&&& 47.82 & 67.39 & 55.85  \\ \cline{4-6}
$\vdots$\\
&&& 48.52 & 68.76 & 56.85  \\ \cline{4-6}
&&& 51.63 & 61.29 & 55.98  \\
&&& 51.31 & 64.10 & 56.94  \\ \cline{4-6}

\end{tabular}

}
\caption{%
Precision, Recall and F1 score for: the base models; the pipeline models; base models trained with an increasing number of \negadeG samples; pipeline models trained with all \negadeG samples.
}
\label{tab:p_r_f1}
\end{table*}

All the reported results are the average over 5 runs. For the Transformer models we used the same hyperparameters reported by \citet{portelli2021bert}.

As metrics, we consider
the number of false positive predictions (FP)
and the relaxed precision (P), recall (R) and F1 score as defined in the \SMM shared tasks \citep{weissenbacher2019overview}: the scores take into account ``partial'' matches, in which it is sufficient for a prediction to partially overlap with the gold annotation.

\smallskip

As a preliminary step, the two negation detection models are trained and used to predict the  negation scopes for all the test samples once.
This allows us to compute the predictions of any pipeline model.

\textbf{Exp 1:}
to provide a measure of the initial robustness of the base models and their general performance, we train
them on the \ade and \noade samples only (\the\numADETrain+\the\numNoADETrain\ samples).
We then test the efficacy of the pipeline negation detection method, applying NegEx and BERTneg 
to the base models.

\textbf{Exp 2:}
to test the effect of augmenting the training data with artificial samples (i.e., the second negation detection method), we add to the training set an increasing number of \negadeG samples (\smalltt{+50} to \smalltt{+\the\numNegADETrain}, in steps of 50 samples). 
During preliminary experiments, we added 100 \noade samples from \SMMib to the training set. The performance of all the models \textit{did not vary} in this case, showing that the results of Exp 2 are caused by the nature of the samples, and not simply by the increased size of training set
All the base models are then fine-tuned on the resulting dataset.

\textbf{Exp 3:} to investigate whether the two methods are complementary in their action, we combine the two strategies, applying the pipeline architecture to the models trained on the augmented dataset.

\begin{figure*}[htbp!]
\centering
\includegraphics[
width=1.\linewidth,
trim={14cm .8cm 62cm 0.5cm},
clip
]{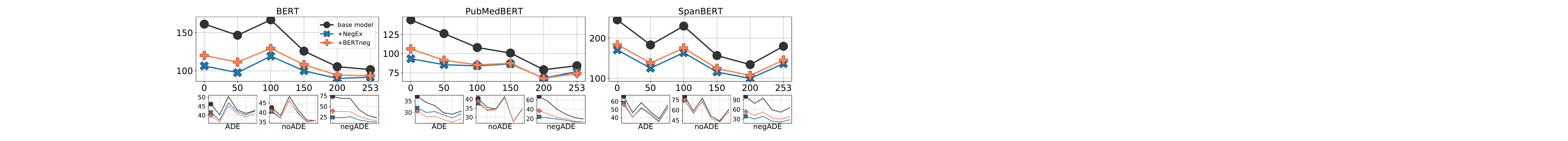}
\caption{
Top: total FP for the three base models and their pipeline versions with the increasing of \negade samples in the training set (x-axis: from 50 to \the\numNegADETrain).
Bottom: number of FP for all the models by sample category.
}
    \label{fig:spurious}
\end{figure*}

\subsection{Results}

Table \ref{tab:fp} and \ref{tab:p_r_f1} contain a summary of the most relevant metrics for the tested models. We report the number of FP both on the whole test set and on individual partitions (\ade, \noade and \negadeR samples).

\textbf{Exp 1 (rows 1--3):}
all base models (row 1) have a high number of FP, especially in the \negade category. This strongly suggests that they are not robust against this phenomenon.
When combined with NegEx (row 2), the FP decreases by 34\%, showing that the regular expression module removes a great number of unwanted predictions.
\bertneg decreases the number of FP, too, but only by 25\%.
This is due to the difference between the brute-force behaviour of NegEx and the contextual prediction of a deep machine learning model.
We can notice that the two pipeline models slightly reduce the number of FP also in the \ade category (e.g. from 66.2 to 56.2 for SpanBERT).

However, if we look at P and R in the first three rows, we can see that the negation detection modules bring an increase in P at the cost of large drops in R.
Some correct predictions of the base models get discarded, in particular the ADEs that contain a negation (e.g., ``After taking this drug \textit{I cannot sleep anymore}''). As this effect is undesirable, we investigated the use of the \negadeG samples to mitigate it.

\textbf{Exp 2 (rows 4--8):}
adding \negadeG samples to the training set (from \smalltt{+50} to \smalltt{+253}) lowers the number of FP predictions for all models.
Using all the available samples brings down the number of FP as much as using NegEx on the initial model (compare {+NegEx} and \smalltt{+253}).
This reduction is given by the decrease of FP in the \negade set, while the number of FP in the \ade and \noade categories remains roughly stable.

Comparing row \smalltt{+253} with row 1 shows, as for Exp 1, an increase in P and a drop in R. However, the drop in R is less severe than before (5 points at most), meaning that less true positives are being discarded. Also, P increases twice as much, leading to an overall increase in F1.

\textbf{Exp 3 (rows 9--10):}
the effect of augmented dataset and negation modules are complementary, as shown by the further decrease in FP.
However, combining the two approaches does not seem to be a winning strategy, as it leads to a further decrease in R without the benefit of increasing P.

The same behaviour can be observed for all the base models, despite the different initial performance (with SpanBERT having a generally higher R and PubMedBERT an higher P).

\medskip

Figure \ref{fig:spurious} offers another visualization of the effect that adding \negadeG samples has on the number of False Positives.
The number of \negade samples in the training set increases from left to right in each plot (from 0 to \the\numNegADETrain).
The top row shows how the total number of predicted FPs decreases (for all base and pipeline models) when adding the generated negated samples.
The plots in the bottom row show how the number of FP varies for the three categories of samples separately (\ade, \noade and \negade, bottom row). As observed in Table \ref{tab:fp}, the decrease is most significant in the \negade partition.

\medskip

The results show that introducing a small number of new samples (even if artificial) is the best way to directly increase the model knowledge about the phenomenon.
However, this solution could be expensive in absence of a large quantity of negated data. For this reason, the pipeline models might be a viable alternative, as they maintain the F1 score while still decreasing the number of false positives.

\section{Conclusions}

In this paper, we evaluate the impact of negations on state-of-the-art ADE detection models.
We introduced \datasetname, a new dataset specifically aimed at studying this phenomenon.
The dataset proves to be a challenging setting and the experiments show that current methods lack mechanisms to deal with negations.
We introduce and compare two strategies to tackle the problem: using a negation detection module and adding \negadeG samples in the training set.
Both of them bring significant increases in performance.

Both the dataset and the code are made publicly available for the community to test the robustness of their systems against negation.

Future work should focus on more refined techniques to accurately model the semantic properties of the samples, also by jointly handling negation and speculation phenomena. This might be an essential requirement for dealing with the noisiness and variety of social media texts. The main short term directions are increasing the quality and quantity of real \negade samples (possibly via crowd-sourcing), and creating a model that is able to discard \negade (keeping an high precision level), without sacrificing recall. 

\section{Acknowledgments}

We would like to thank the three anonymous reviewers for their insightful feedback.

\bibliography{anthology,custom}
\bibliographystyle{acl_natbib}

\appendix

\section{Data Augmentation Process}
\label{appendix:data}
The data augmentation process was carried out by four volunteers with a high level of proficiency in English.
More specifically, the volunteers were: two graduate students (Master's degree in Computer Science and Artificial Intelligence) and two Ph.D. in Natural Language Processing. All of them have a minimum English level of C1.\footnote{\url{https://www.efset.org/cefr/c1/}}

\subsection{Recovery of Real Samples}
The \noade samples from the \SMMib and \SMMiib binary classification datasets have been filtered using the negation cues from BioScope (e.g. \textit{none}, \textit{missing}, \textit{no longer}, \textit{etc.}). Thanks to this first filtering, only the remaining 3897 samples have been analyzed by the volunteers. A tweet was kept only if it negated the presence of an ADE.

In the following examples, 1 and 2 are valid tweets, while 3 and 4 contain negations but do not negate the ADE:

\begin{enumerate}
\setlength\itemsep{0mm}
\item 
\smallsf{This \textbf{\#HUMIRA shot} has me feeling like a normal human... \textbf{No pain no inflammation} no nothinggggh \#RAproblems}

\item
\smallsf{@UKingsbrook That's correct! \textbf{Metoprolol} is \textbf{NOT known to cause hypokalemia}.}

\item
\smallsf{I've seen so much \textbf{Tamiflu} these past couple of days I'm \textbf{not even surprised I'm shivering and experiencing aches} right now. *sigh}

\item
\smallsf{But I'm \textbf{not on adderall and I am feasting}.}
\end{enumerate}

\subsection{Generation of Artificial Samples}
According to the split provided by Portelli et al. (2021), we extracted and modified only the \ade samples from the test set (260 samples). The volunteers were instructed to modify the samples with as little edits as possible, while still generating a plausible tweet. They were encouraged to add one or more negation cue words to negate the ADE reported in the tweet. In the case it was not possible to negate the meaning of the tweet just by adding cue words, they were allowed to perform more edits in the sentence and use longer expressions.

Each annotator was asked to review the edits done by the others, and asked to point out which samples seemed unrealistic or failed to negate the ADE.
During the augmentation process, if no agreement was found about the edits, the tweet modified tweet was discarded. At the end of the process 7 tweets were removed from the final dataset.
Due to the generation process we implemented, we could not directly measure the inter-annotator agreement, which could, however, be inferred by the number of discarded samples.

\section{Full Results}
\label{appendix:results}

Table \ref{tab:full_results} reports the metrics for all the models (average over 5 runs).
The table includes also the results for the pipeline models trained with an increasing number of \negadeG samples, which show the same trend as the base models.
The most relevant combinations (discussed in the main part of the paper) are highlighted in color (%
base models in gray{\color{gray!75}$\blacksquare$}, 
base models trained with 253 \negadeG samples in orange{\color{orange!40}$\blacksquare$},
pipeline models in blue{\color{teal!30}$\blacksquare$}%
).

\begin{table*}[!ht]
\centering
\resizebox{.49\linewidth}{!}{%
\begin{tabular}{lrrrrrr}
& \multicolumn{6}{c}{\Large \textbf{False Positives}} \\ \hline
\textbf{\negadeG samples} &\multicolumn{1}{c}{\makebox[1cm][c]{\textbf{0}}} &
\multicolumn{1}{c}{\makebox[1cm][c]{\textbf{50}}} &
\multicolumn{1}{c}{\makebox[1cm][c]{\textbf{100}}} &
\multicolumn{1}{c}{\makebox[1cm][c]{\textbf{150}}} &
\multicolumn{1}{c}{\makebox[1cm][c]{\textbf{200}}} &
\multicolumn{1}{c}{\makebox[1cm][c]{\textbf{253}}} \\
 \hline
{BERT}& \cellred 161.20 & 146.60 & 166.80 & 125.80 & 105.40 & \cellblue 101.60 \\
{\ \ +NegEx}& \cellgreen 106.40 & 97.60 & 119.20 & 100.00 & \textbf{90.00} & 91.60 \\
{\ \ +BERTneg}& \cellgreen 120.20 & 111.40 & 129.40 & 108.00 & 94.40 & 93.60 \\ \hline
{PubMedBERT}& \cellred 144.20 & 126.20 & 108.00 & 100.80 & 79.00 & \cellblue 84.20 \\
{\ \ +NegEx}& \cellgreen 93.40 & 85.60 & 84.00 & 86.60 & 68.00 & 76.60 \\
{\ \ +BERTneg}& \cellgreen 106.20 & 91.40 & 85.40 & 87.40 & \textbf{67.60} & 74.20 \\ \hline
{SpanBERT}& \cellred 245.60 & 183.20 & 230.00 & 156.80 & 134.20 & \cellblue 179.80 \\
{\ \ +NegEx}& \cellgreen 170.80 & 125.40 & 163.80 & 115.80 & \textbf{99.80} & 136.40 \\
{\ \ +BERTneg}& \cellgreen 184.00 & 138.40 & 175.40 & 124.60 & 107.20 & 145.40 \\
\hline
\end{tabular}}\hfill
\resizebox{.49\linewidth}{!}{%
\begin{tabular}{lrrrrrr}
& \multicolumn{6}{c}{\Large \textbf{Precision}} \\ \hline
\textbf{\negadeG samples} &\multicolumn{1}{c}{\makebox[1cm][c]{\textbf{0}}} &
\multicolumn{1}{c}{\makebox[1cm][c]{\textbf{50}}} &
\multicolumn{1}{c}{\makebox[1cm][c]{\textbf{100}}} &
\multicolumn{1}{c}{\makebox[1cm][c]{\textbf{150}}} &
\multicolumn{1}{c}{\makebox[1cm][c]{\textbf{200}}} &
\multicolumn{1}{c}{\makebox[1cm][c]{\textbf{253}}} \\
 \hline
{BERT}& \cellred 50.15 & 51.47 & 49.17 & 54.24 & 57.82 & \cellblue 58.85 \\
{\ \ +NegEx}& \cellgreen 55.59 & 56.79 & 53.52 & 56.53 & 58.52 & 58.65 \\
{\ \ +BERTneg}& \cellgreen 54.37 & 55.34 & 52.78 & 55.88 & 58.53 & \textbf{59.22} \\ \hline
{PubMedBERT}& \cellred 53.24 & 56.43 & 59.25 & 59.46 & 62.94 & \cellblue 63.28 \\
{\ \ +NegEx}& \cellgreen 59.21 & 61.60 & 61.98 & 60.29 & 63.84 & 63.24 \\
{\ \ +BERTneg}& \cellgreen 57.69 & 61.24 & 62.44 & 61.12 & \textbf{64.84} & 64.74 \\ \hline
{SpanBERT}& \cellred 43.65 & 53.32 & 45.46 & 48.24 & 52.80 & \cellblue 48.52 \\
{\ \ +NegEx}& \cellgreen 48.65 & \textbf{60.44} & 49.95 & 52.83 & 55.95 & 51.63 \\
{\ \ +BERTneg}& \cellgreen 47.82 & 56.60 & 49.25 & 51.43 & 55.54 & 51.31 \\
\hline
\end{tabular}}\medskip

\resizebox{.49\linewidth}{!}{%
\begin{tabular}{lrrrrrr}
& \multicolumn{6}{c}{\Large \textbf{F1 score}} \\ \hline
\textbf{\negadeG samples} &\multicolumn{1}{c}{\makebox[1cm][c]{\textbf{0}}} &
\multicolumn{1}{c}{\makebox[1cm][c]{\textbf{50}}} &
\multicolumn{1}{c}{\makebox[1cm][c]{\textbf{100}}} &
\multicolumn{1}{c}{\makebox[1cm][c]{\textbf{150}}} &
\multicolumn{1}{c}{\makebox[1cm][c]{\textbf{200}}} &
\multicolumn{1}{c}{\makebox[1cm][c]{\textbf{253}}} \\
 \hline
{BERT}& \cellred 56.78 & 56.59 & 56.18 & 58.41 & 59.45 & \cellblue \textbf{61.21} \\
{\ \ +NegEx}& \cellgreen 56.73 & 56.48 & 56.10 & 57.03 & 57.19 & 58.29 \\
{\ \ +BERTneg}& \cellgreen 57.30 & 56.73 & 56.61 & 57.65 & 58.06 & 59.65 \\ \hline
{PubMedBERT}& \cellred 59.47 & 61.37 & 62.72 & 61.59 & 61.74 & \cellblue \textbf{63.20} \\
{\ \ +NegEx}& \cellgreen 59.47 & 60.91 & 61.26 & 59.22 & 59.26 & 60.58 \\
{\ \ +BERTneg}& \cellgreen 60.04 & 62.15 & 62.70 & 61.10 & 61.36 & 62.72 \\ \hline
{SpanBERT}& \cellred 54.61 & 45.76 & 56.20 & 49.47 & 54.75 & \cellblue 56.85 \\
{\ \ +NegEx}& \cellgreen 55.55 & 46.54 & 56.69 & 48.85 & 53.44 & 55.98 \\
{\ \ +BERTneg}& \cellgreen 55.85 & 46.61 & \textbf{57.17} & 49.61 & 54.49 & 56.94 \\
\hline
\end{tabular}}\hfill
\resizebox{.49\linewidth}{!}{%
\begin{tabular}{lrrrrrr}
& \multicolumn{6}{c}{\Large \textbf{Recall}} \\ \hline
\textbf{\negadeG samples} &\multicolumn{1}{c}{\makebox[1cm][c]{\textbf{0}}} &
\multicolumn{1}{c}{\makebox[1cm][c]{\textbf{50}}} &
\multicolumn{1}{c}{\makebox[1cm][c]{\textbf{100}}} &
\multicolumn{1}{c}{\makebox[1cm][c]{\textbf{150}}} &
\multicolumn{1}{c}{\makebox[1cm][c]{\textbf{200}}} &
\multicolumn{1}{c}{\makebox[1cm][c]{\textbf{253}}} \\
 \hline
{BERT}& \cellred \textbf{65.60} & 63.04 & 65.57 & 63.41 & 61.85 & \cellblue 63.86 \\
{\ \ +NegEx}& \cellgreen 58.03 & 56.32 & 58.97 & 57.66 & 56.43 & 58.03 \\
{\ \ +BERTneg}& \cellgreen 60.70 & 58.38 & 61.09 & 59.66 & 58.16 & 60.17 \\ \hline
{PubMedBERT}& \cellred 67.41 & \textbf{67.44} & 66.67 & 64.03 & 60.67 & \cellblue 63.33 \\
{\ \ +NegEx}& \cellgreen 59.76 & 60.33 & 60.59 & 58.32 & 55.45 & 58.29 \\
{\ \ +BERTneg}& \cellgreen 62.64 & 63.19 & 62.99 & 61.19 & 58.36 & 60.98 \\ \hline
{SpanBERT}& \cellred 73.28 & 59.73 & \textbf{74.19} & 59.22 & 60.32 & \cellblue 68.76 \\
{\ \ +NegEx}& \cellgreen 65.04 & 53.01 & 65.95 & 52.86 & 53.91 & 61.29 \\
{\ \ +BERTneg}& \cellgreen 67.39 & 54.98 & 68.48 & 55.14 & 56.27 & 64.10 \\
\hline
\end{tabular}}

\caption{All metrics for: the original base models (gray{\color{gray!75}$\blacksquare$});  the base models trained with different quantities of \negadeG samples (the results of training with 253 \negadeG samples in orange{\color{orange!40}$\blacksquare$}); the two pipeline models (in blue{\color{teal!30}$\blacksquare$}); the combination of the two methods. For each model, in bold is highlighted the best value for the specific evaluation metric.}
\label{tab:full_results}
\end{table*}

\end{document}